\documentclass{article}
\usepackage{spconf, IEEEtrantools}
\usepackage{cite, enumitem}
\usepackage{amsmath,amssymb,amsfonts}
\usepackage{algorithm}
\usepackage{algorithmic}
\usepackage{hyperref}
\usepackage{graphicx}
\usepackage{textcomp}   
\usepackage{textcomp}
\usepackage{xcolor}
\usepackage[normalem]{ulem} 
\usepackage{multicol, multirow}
\usepackage{setspace}
\AtBeginEnvironment{thebibliography}{\footnotesize}

\newcommand{\round}[1]{\ensuremath{\left\lfloor#1\right\rceil}}

\title{EntroLLM: \underline{Entro}py Encoded Weight Compression for Efficient \underline{L}arge \underline{L}anguage \underline{M}odel Inference on Edge Devices \vspace{-3mm} \\} 
\name{Arnab Sanyal\IEEEauthorrefmark{1}\IEEEauthorrefmark{2}, Gourav Datta\IEEEauthorrefmark{4}, Prithwish Mukherjee\IEEEauthorrefmark{3}, Sandeep P. Chinchali\IEEEauthorrefmark{2}, Michael Orshansky\IEEEauthorrefmark{2}
\vspace{-3mm}
\thanks{\IEEEauthorrefmark{1} Corresponding Author -- \href{mailto:sanyal@utexas.edu}{sanyal@utexas.edu}\\
		Authors would like to thank \href{mailto:amirgh@berkeley.edu}{Amir Gholami}, \href{mailto:chooper@berkeley.edu}{Coleman Richard Charles Hooper}, and \href{mailto:keutzer@berkeley.edu}{Kurt Keutzer} for their valuable input during ideation.}
}
	
\address{
	\begin{minipage}[t]{0.30\textwidth}
		\centering
		\IEEEauthorrefmark{2}The University of Texas at Austin
	\end{minipage}
	\begin{minipage}[t]{0.30\textwidth}
		\centering
		\IEEEauthorrefmark{3}Georgia Institute of Technology
	\end{minipage}
	\begin{minipage}[t]{0.30\textwidth}
		\centering
		\IEEEauthorrefmark{4}Case Western Reserve University
	\end{minipage}
    \vspace{-5mm}
}
\begin{document}

%
\maketitle
\begin{abstract}
\vspace{-1mm}
Large Language Models (LLMs) achieve strong performance across tasks, but face storage and compute challenges on edge devices. We propose EntroLLM, a compression framework combining mixed quantization and entropy coding to reduce storage while preserving accuracy. We use a combination of unsigned and asymmetric quantization. Tensor-level quantization produces an entropy-reducing effect, increasing weight compressibility, and improving downstream Huffman encoding by $7\times$ (8-bit) and $11.3\times$ (4-bit) over state-of-the-art methods. Huffman coding further reduces memory bandwidth demands, while a parallel decoding strategy enables efficient weight retrieval with minimal latency. Experiments on edge-scale LLMs (\textit{smolLM-1.7B}, \textit{phi3-mini-4k}, \textit{mistral-7B}) show up to 30\% storage savings over \texttt{uint8} and 65\% over \texttt{uint4} models, with 31.9–146.6\% faster inference on memory-limited devices like the \textsc{NVIDIA Jetson P3450}. EntroLLM requires no retraining and is compatible with existing post-training quantization pipelines, making it practical for edge LLM deployment.
\end{abstract}
\vspace{-4mm}
\section{Introduction}
\vspace{-2mm}

Large language models (LLMs) have demonstrated remarkable performance in various domains \cite{NEURIPS2020_1457c0d6, Touvron2023LLaMAOA}, but their substantial size poses challenges for deployment, especially on resource-constrained edge devices \cite{zhou2024surveyefficientinferencelarge}. For example, even a smaller LLM such as \textit{mistral-7B-Instruct} \cite{jiang2023mistral7b} requires more than $14$ GB of memory with weights encoded in $16$-bit floats (\texttt{fp16}), exceeding the capacity of most edge devices. As a direct result of being massively parameter-heavy, the primary bottleneck in LLM inference, particularly for generative tasks, is memory bandwidth rather than computation \cite{kim2023squeezellm}. This means that the speed of loading and storing parameters becomes the most critical constraint \cite{memory_wall}.
\begin{figure}[b!]
	\centering
    \vspace{-4mm}
	\includegraphics[width=\columnwidth]{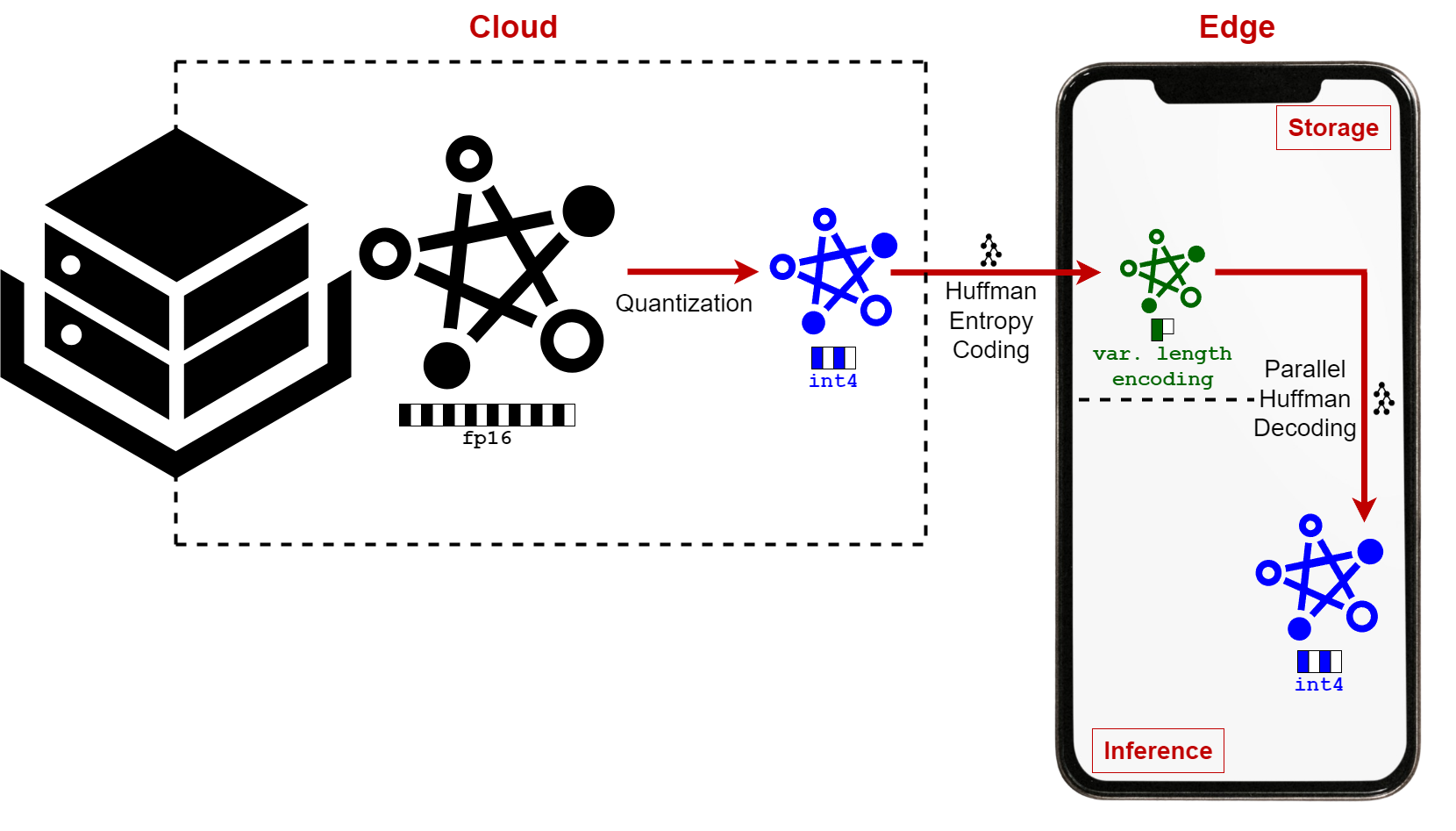}
    \vspace{-8mm}
	\caption{\footnotesize Overall schematic of our edge-device inference scheme. A floating-point model is first quantized using a mixed scheme to maximize compressibility. For storage on edge devices, integer weights are entropy-coded, then decoded in parallel during inference to minimize latency. Section \ref{sec:method} details these components.}
	\label{fig:schematic}
    \vspace{-3mm}
\end{figure}

\begin{figure*}[t!]
	\begin{minipage}[b]{0.32\linewidth}
		\centering
		\centerline{\includegraphics[width=\linewidth]{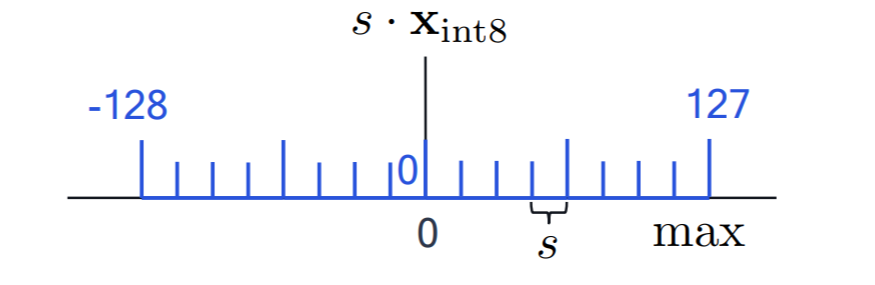}}
		\begin{center}
			(a) \texttt{Symmetric Signed}
		\end{center}
	\end{minipage} \hfill
	\begin{minipage}[b]{0.32\linewidth}
		\centering
		\centerline{\includegraphics[width=\linewidth]{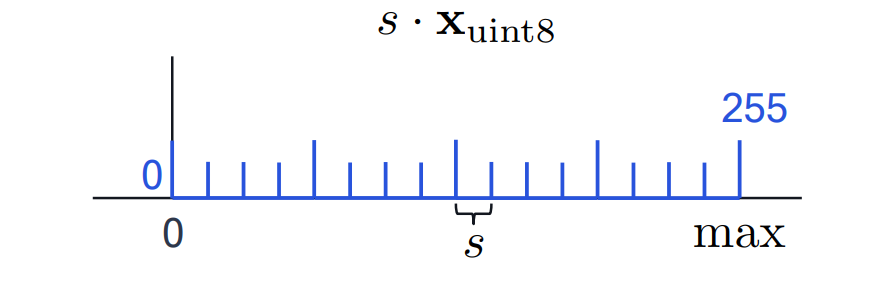}}
		\begin{center}
			(b) \texttt{Unsigned}
		\end{center}
	\end{minipage} \hfill
	\begin{minipage}[b]{0.32\linewidth}
		\centering
		\centerline{\includegraphics[width=\linewidth]{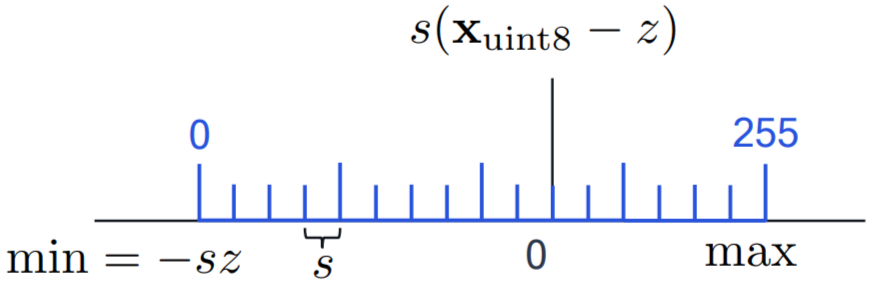}}
		\begin{center}
			(c) \texttt{Asymmetric}
		\end{center}
	\end{minipage}
    \vspace{-3mm}
	\caption{\footnotesize A visual explanation of the different uniform quantization grids \cite{nagel2021whitepaperneuralnetwork} for a bit-width of $8$. $s$ is the scaling factor, $z$ the zero-point. The floating-point grid is in black, and the integer quantized grid is in blue. In our work, we use either an unsigned or an asymmetric quantization scheme on each layer based on the individual layer's weight distribution.}
	\label{fig:quantization}
    \vspace{-5mm}
\end{figure*}

Quantization, a technique that encodes the model parameters in lower precision formats, has effectively reduced the size of LLMs \cite{NEURIPS2022_adf7fa39}. However, further compression without compromising performance remains challenging, especially for smaller LLMs ($\leq$ 10B parameters) designed for edge applications. To address this, we focus on reducing the storage overhead for deploying these small LLMs on low-power edge devices. We introduce an entropy-based encoding scheme to further compress these models without loss of information. Specifically, we leverage variable-length bit widths to encode quantized weights, using techniques rooted in entropy coding to achieve lossless compression. During inference, the original weights are recovered with precision without additional storage overhead, ensuring that performance degradation is minimized to minor quantization effects. This approach, illustrated in Figure \ref{fig:schematic}, employs Huffman coding due to its alignment with the Shannon entropy bounds, offering optimal compression for the quantized model parameters while maintaining the computational efficiency required for edge inference. We also use a tensor-level mixed quantization scheme, comprising asymmetric quantization and unsigned quantization to induce better weight compression, and implement a partitioning scheme for model weights so that parallel decoding is made possible during inference. Our main contributions to this work are as follows --
\begin{itemize}[leftmargin=*]
\vspace{-2.7mm}
	\item \underline{\textit{Mixed Quantization Scheme}}: We introduce a layer-specific quantization strategy that balances storage efficiency with model performance. This approach significantly improves the downstream compressibility of entropy compared to state-of-the-art quantization methods (as shown in table \ref{tab:compression}).
    \vspace{-6mm}
	\item \underline{\textit{Huffman Weight Encoding}}: We leverage entropy-based Huffman coding to compress quantized weights further, reducing storage overhead while preserving accuracy.
    \vspace{-2.7mm}
	\item \underline{\textit{Parallel Decoding for Efficient Inference}}: To mitigate latency issues caused by sequential entropy decoding, we implement a parallelized decoding strategy that enables real-time processing on edge hardware.
    \vspace{-2.7mm}
	\item \underline{\textit{Comprehensive Evaluation}}: We evaluated our approach on multiple LLMs (\textit{smolLM-1.7B-Instruct} \cite{SmolLM-2024}, \textit{phi3-mini-4k-Instruct} \cite{phi3-2024}, and \textit{mistral-7B-Instruct} \cite{jiang2023mistral7b}) deployed on devices with limited resources, demonstrating significant storage savings and reduced inference latency.
\end{itemize}

\vspace{-6mm}
\section{Related Works}
\label{sec:rel_works}
\vspace{-2mm}

Before LLMs, compression techniques such as pruning \cite{han2015learning}, quantization \cite{quantization, banner2019post}, knowledge distillation \cite{distillation}, and alternative number formats such as logarithmic \cite{lnsdnn} and posit \cite{posit} were developed to improve deep learning efficiency. These typically require retraining, which is impractical for LLMs due to extreme memory demands. Recent work focuses instead on LLM compression with minimal or no retraining.

\noindent\textbf{Pruning}: Sun et al. \cite{sun2024a} proposed an approach that prunes weights with the smallest magnitudes multiplied by the corresponding input activations per output. LLM-pruner \cite{ma2023llmpruner} adopted structural pruning that selectively removes non-critical coupled structures based on gradient information, maximally preserving the main functionality of the LLM.

\noindent\textbf{Quantization}: Post-training quantization methods such as GPTQ \cite{frantar2023optq}, AWQ \cite{lin2023awq}, and SpQR \cite{dettmers2023spqrsparsequantizedrepresentationnearlossless} reduce LLM memory by fine-tuning weights with small calibration datasets, allowing $4$-bit precision with minimal accuracy loss. Although most focus on uniform quantization, sensitivity-based non-uniform approaches \cite{dettmers2023qlora, kim2023squeezellm} allow for more aggressive quantization. Ultra-low-precision models, including ternary LLMs \cite{bondarenko-etal-2021-understanding,wei2022outlier,chen2024ternaryllmternarizedlargelanguage}, address activation and weight outliers. However, their use for small edge-focused LLMs remains largely unexplored. 


\vspace{-4mm}
\section{Proposed Method}
\label{sec:method}
\vspace{-3mm}

Entropy coding compresses the data by exploiting symbol frequencies. Huffman encoding \cite{huffenc} optimally assigns shorter codes to frequent symbols and is the core of the beyond-quantization compression framework in our work. \\
\noindent\textbf{Mixed Quantization Scheme}: We aim to compress weights beyond post-training quantization. State-of-the-art models often utilize block-based quantization (e.g., blocks of 32 weights), which flattens the integer set and increases entropy. In contrast, obtaining a set of integer codes for an entire tensor natively results in a much ``spikier'' distribution with lower entropy. To preserve accuracy while leveraging this tensor-level compressibility, we employ a mixed quantization scheme, applying either unsigned (left of Eq. \ref{eq:assym}) or asymmetric (right of Eq. \ref{eq:assym}) quantization per layer, based on its weight distribution. Fig. \ref{fig:quantization} illustrates these schemes and their mapping from floating-point to integer grids.
\vspace{-2mm}
\begin{align}
	\mathtt{W_{int}} \leftarrow \round{\frac{\mathtt{W_{fp}}}{s}}, \ \ \ 
	\mathtt{W_{int}} \leftarrow \round{\frac{(\mathtt{W_{fp}} - z)}{s}} \label{eq:assym}
\end{align}
\vspace{-2mm}

\begin{figure}[t!]
	\centering
	\includegraphics[width=\columnwidth]{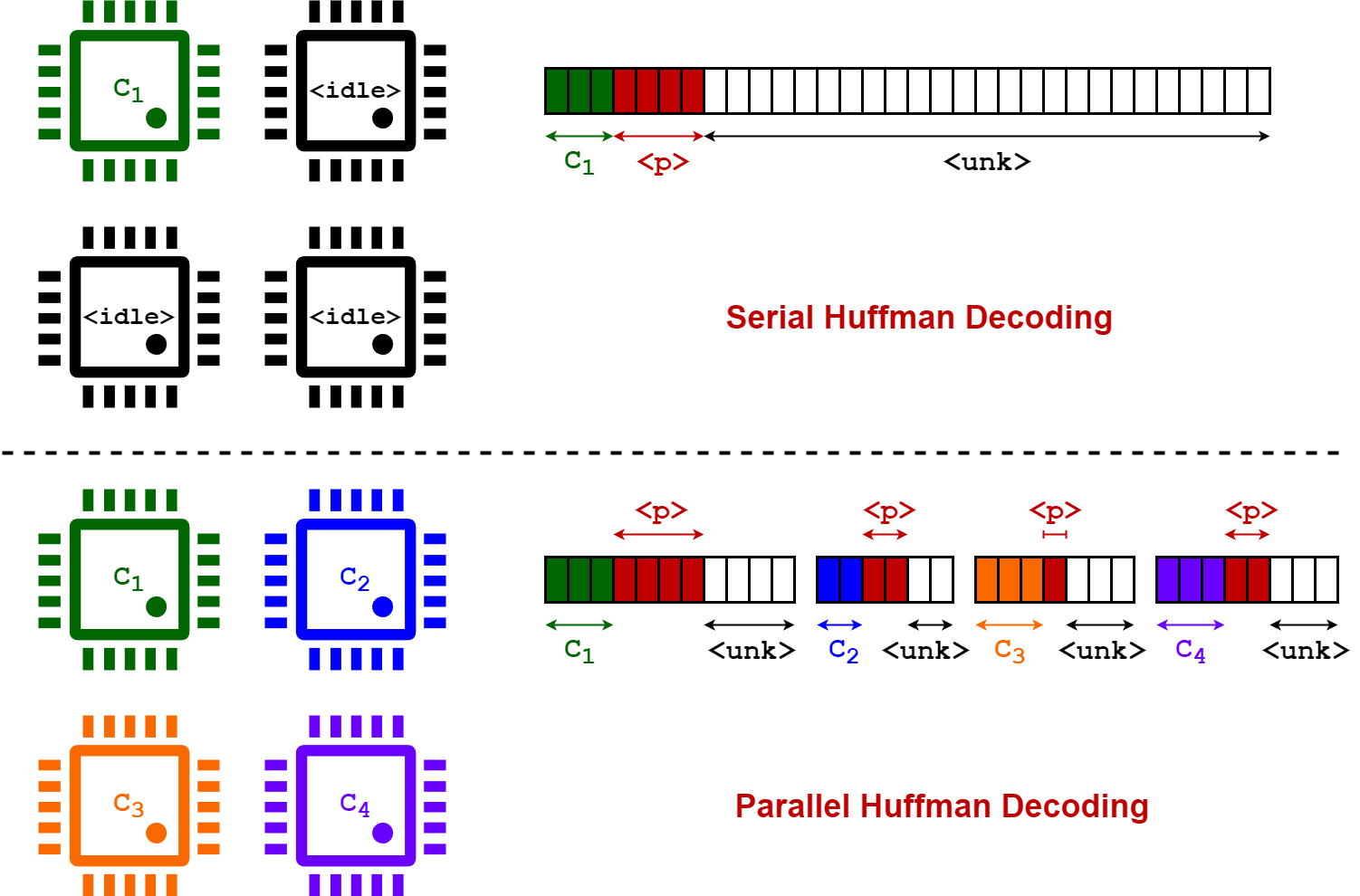}
    \vspace{-5mm}
	\caption{\footnotesize In the serial decoding setting, parallelization is not possible as the variable code lengths make it difficult for us to predict the start token of a symbol in our remaining encoded parameters. By keeping the parameter space's original weight tensor packing structure intact, we can parallelize decoding, as now we can assign different encoded tensors (chunks) to individual threads assigned to cores. Modern LLMs have hundreds to thousands of such weight tensors, and hence, coarse-grained parallelism is achievable. In the figure {\color{red}\texttt{<p>}} represents current bits being decoded, while \texttt{<unk>} represents unknown encoded parameter bits.}
	\label{fig:parallel-decoding}
     \vspace{-6mm}
\end{figure}
\vspace{-2mm}
By quantizing at the tensor level, we maintain a ``spikier'' distribution of integer codes compared to block-based approaches. This spikiness directly translates into lower entropy and higher compressibility. Using the low entropy produced by this tensor-level strategy, we generate an efficient Huffman tree, allowing us to encode the model weights with minimal redundancy while maintaining precision during inference. This synergy between the quantization strategy and entropy coding not only compresses the model effectively but also ensures that the performance remains unaffected by the transformation. As shown in Table \ref{tab:compression}, our quantization scheme improves downstream entropy compressibility by a factor of up to $\mathbf{8.1\times}$ for $8$-bit models and $\mathbf{13.1\times}$ for $4$-bit models compared to SOTA techniques. 

\begin{figure*}[t!]
	\begin{minipage}[b]{0.32\linewidth}
		\centering
	\centerline{\includegraphics[width=\linewidth]{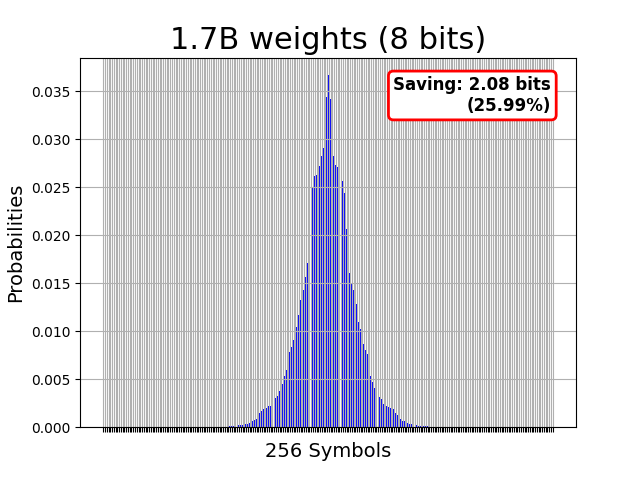}}
    \vspace{-2mm}
		\begin{center}
			(a) \texttt{smolLM-1.7B-Instruct}
		\end{center}
        \vspace{-6mm}
		\medskip
	\end{minipage}
	\hfill
	\begin{minipage}[b]{0.32\linewidth}
		\centering
		\centerline{\includegraphics[width=\linewidth]{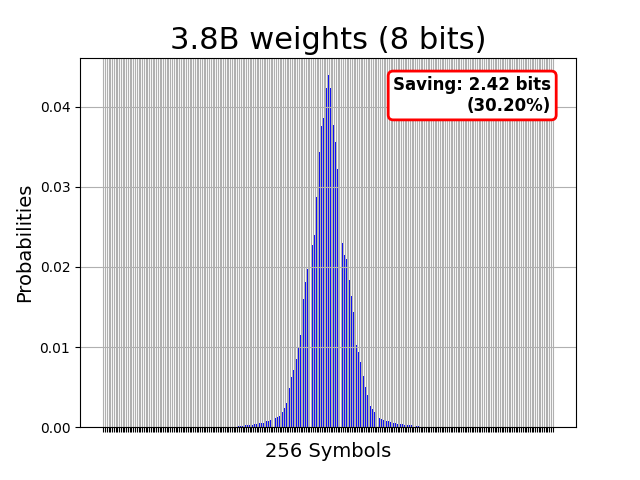}}
        \vspace{-2mm}
		\begin{center}
			(b) \texttt{phi3-mini-4k-Instruct}
		\end{center}
        \vspace{-6mm}
		\medskip
	\end{minipage}
	\hfill
	\begin{minipage}[b]{0.32\linewidth}
		\centering
		\centerline{\includegraphics[width=\linewidth]{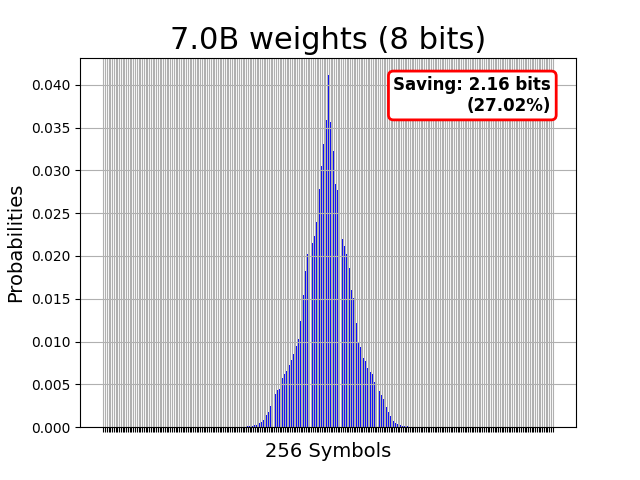}}
        \vspace{-2mm}
		\begin{center}
			(c) \texttt{mistral-7B-Instruct}
		\end{center}
        \vspace{-6mm}
		\medskip
	\end{minipage}
	\hfill
	\begin{minipage}[b]{0.32\linewidth}
		\centering
		\centerline{\includegraphics[width=\linewidth]{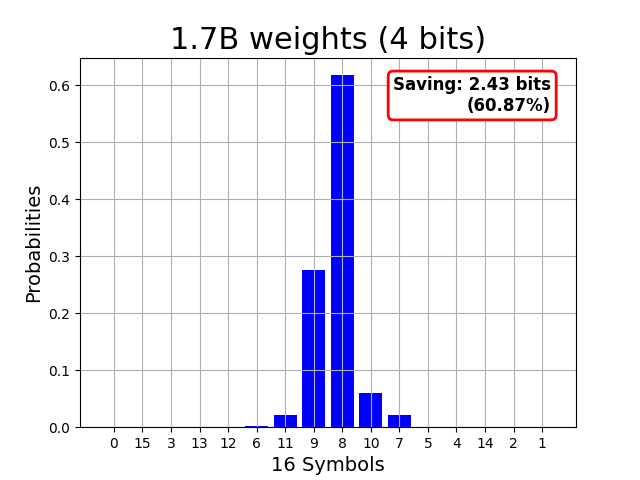}}
		\begin{center}
			(d) \texttt{smolLM-1.7B-Instruct}
		\end{center}
		\medskip
	\end{minipage}
	\hfill
	\begin{minipage}[b]{0.32\linewidth}
		\centering
		\centerline{\includegraphics[width=\linewidth]{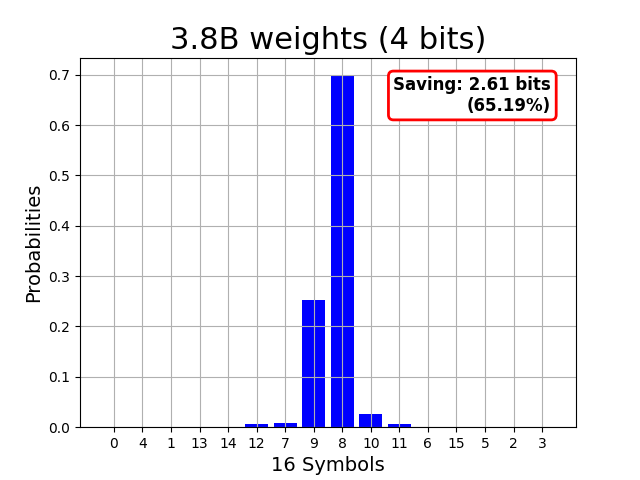}}
		\begin{center}
			(e) \texttt{phi3-mini-4k-Instruct}
		\end{center}
		\medskip
	\end{minipage}
	\hfill
	\begin{minipage}[b]{0.32\linewidth}
		\centering
		\centerline{\includegraphics[width=\linewidth]{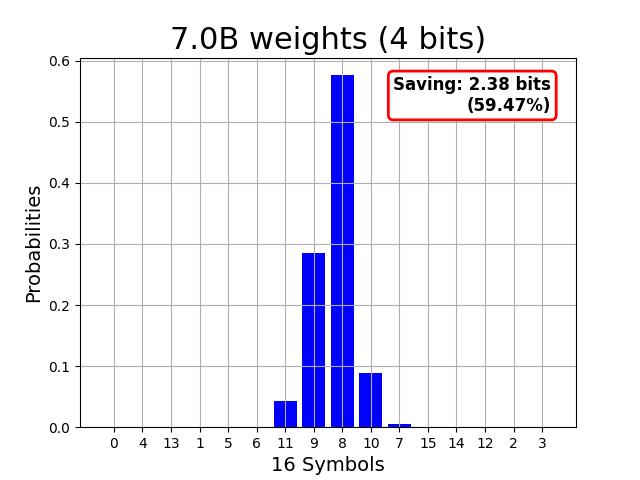}}
		\begin{center}
			(f) \texttt{mistral-7B-Instruct}
		\end{center}
		\medskip
	\end{minipage}
    \vspace{-8mm}
	\caption{\footnotesize \textbf{Top:} Parameter distribution for 8-bit quantized models. \textbf{Bottom:} Parameter distribution for 4-bit counterparts. Model size increases from left to right.}
	\label{fig:plots}
\end{figure*}

\begin{table*}[ht!]
	\centering
	\resizebox{\textwidth}{!}{ 
		\begin{tabular}{||c||c|c|c|c|c|c|c|c|c||}
			\hline\hline
			\multirow{2}{*}{\textsc{Property}} & \multicolumn{9}{c||}{\textsc{Models}} \\ \cline{2-10}
			& \multicolumn{3}{c|}{\texttt{smolLM-Instruct}} & \multicolumn{3}{c|}{\texttt{phi3-mini-4k-Instruct}} & \multicolumn{3}{c||}{\texttt{mistral-Instruct}} \\
			\hline\hline
			Parameter Count & \multicolumn{3}{c|}{$1.7$ Billion} & \multicolumn{3}{c|}{$3.8$ Billion} & \multicolumn{3}{c||}{$7.0$ Billion} \\ \hline
			\multirow{2}{*}{Weight Encoding} & \multirow{2}{*}{\texttt{fp16}} & \texttt{uint8} & \texttt{uint4} & \multirow{2}{*}{\texttt{fp16}} & \texttt{uint8} & \texttt{uint4} & \multirow{2}{*}{\texttt{fp16}} & \texttt{uint8} & \texttt{uint4}  \\ 
            & & (SOTA) & (SOTA) & & (SOTA) & (SOTA) & & (SOTA) & (SOTA) \\ \hline
			\textsc{Wikitext2} (ppl.) $\downarrow$ & $\mathbf{23.81}$ & $23.92$ & $24.15$ & $\mathbf{9.03}$ & $9.45$ & $10.11$ & $\mathbf{8.17}$ & $8.22$ & $8.28$ \\ \hline
			\textsc{HellaSwag} (acc.) $\uparrow$ & \textbf{25.87\%} & $25.55\%$ & $25.28\%$ & \textbf{82.2\%} & $82.11\%$ & $81.00\%$ & \textbf{58.37\%} & $58.34\%$ & 58.23\% \\ \hline
			\textsc{GSM8k CoT} (acc.) $\uparrow$ & - & - & - & \textbf{77.37\%} & $72.85\%$ & $70.57\%$ & \textbf{52.2\%} & $48.63\%$ & $45.38\%$ \\
			\hline\hline
            \multirow{2}{*}{Weight Encoding} & \multirow{2}{*}{-} & \texttt{uint8} & \texttt{uint4} & \multirow{2}{*}{-} & \texttt{uint8} & \texttt{uint4} & \multirow{2}{*}{-} & \texttt{uint8} & \texttt{uint4}  \\ 
            & & (Ours) & (Ours) & & (Ours) & (Ours) & & (Ours) & (Ours) \\ \hline
			\textsc{Wikitext2} (ppl.) $\downarrow$ & - & $23.93$ & $24.14$ & - & $9.44$ & $10.10$ & - & $8.24$ & $8.29$ \\ \hline
			\textsc{HellaSwag} (acc.) $\uparrow$ & - & $25.55\%$ & $25.30\%$ & - & $82.10\%$ & $81.01\%$ & - & $58.33\%$ & 58.21\% \\ \hline
			\textsc{GSM8k CoT} (acc.) $\uparrow$ & - & - & - & - & $72.84\%$ & $70.58\%$ & - & $48.62\%$ & $45.36\%$ \\
			\hline\hline
	\end{tabular}}
    \vspace{-7mm}
	\medskip
	\caption{\footnotesize \textbf{Benchmarks:} Perplexity \& Accuracy for \textit{smolLM-1.7B-Instruct}, \textit{phi3-mini-4k-Instruct} and \textit{mistral-7B-Instruct} on various language tasks}
	\label{tab:pplbenchmark}
    \vspace{-2mm}
\end{table*}

\begin{table*}[ht!]
	\centering
	\resizebox{\textwidth}{!}{ 
		\begin{tabular}{||c||c|c|c|c|c|c|c|c|c||}
			\hline\hline
			\multirow{2}{*}{\textsc{Property}} & \multicolumn{9}{c||}{\textsc{Models}} \\ \cline{2-10}
			& \multicolumn{3}{c|}{\texttt{smolLM-Instruct}} & \multicolumn{3}{c|}{\texttt{phi3-mini-4k-Instruct}} & \multicolumn{3}{c||}{\texttt{mistral-Instruct}} \\
			\hline\hline
			Quantization Type & \multicolumn{3}{c|}{$8$ bits} & \multicolumn{3}{c|}{$8$ bits} & \multicolumn{3}{c||}{$8$ bits} \\ \hline
			Weight Compressibility & SOTA & Ours & Improvement & SOTA & Ours & Improvement & SOTA & Ours & Improvement  \\ \hline
			Bits Saved & $0.29$ & $\mathbf{2.08}$ & $\times7.2$ & $0.30$ & $\mathbf{2.42}$ & $\times8.1$ & $0.31$ & $\mathbf{2.16}$ & $\times7.0$  \\ \hline\hline
			Quantization Type & \multicolumn{3}{c|}{$4$ bits} & \multicolumn{3}{c|}{$4$ bits} & \multicolumn{3}{c||}{$4$ bits} \\ \hline
			Weight Compressibility & SOTA & Ours & Improvement & SOTA & Ours & Improvement & SOTA & Ours & Improvement  \\ \hline
			Bits Saved & $0.21$ & $\mathbf{2.43}$ & $\times11.6$ & $0.20$ & $\mathbf{2.61}$ & $\times13.1$ & $0.21$ & $\mathbf{2.38}$ & $\times11.3$  \\ \hline\hline
	\end{tabular}}
    \vspace{-6mm}
	\medskip
	\caption{\footnotesize A comparison showing how our quantization scheme improves downstream entropy compressibility of weights versus SOTA quantization techniques}
	\label{tab:compression}
    \vspace{-2mm}
\end{table*}

\noindent\textbf{Huffman weight encoding}: Huffman encoding -- which originates from entropy coding -- is a popular data compression technique. The technique is based on the construction of a binary tree, where each symbol is assigned a score corresponding to its frequency of occurrence. This process involves recursively combining pairs of symbols into a tree structure. More frequent symbols are positioned closer to the root of the tree, resulting in shorter paths and consequently shorter binary codes for these symbols. The length of each symbol's binary code is proportional to its distance from the root, thereby minimizing the total length of the encoded data. This ensures that no other prefix code can offer a more efficient representation of the data, making the Huffman encoding optimal for lossless compression. In this work, we employ Huffman encoding to compress the weights of quantized LLMs, enabling their deployment on edge devices. By applying Huffman encoding to the quantized weights of an LLM, particularly those with skewed value distributions, we achieve significant compression without sacrificing accuracy. As discussed in Section \ref{sec:rel_works}, a post-training quantized LLM can match the performance of a full-precision LLM for most language generation benchmarks. Huffman coding converts the model's weight parameters into variable-length codes, assigning shorter codes to frequent weight values, which reduces both storage and bandwidth costs. 

\noindent\textbf{{Segmentation for parallel Huffman decoding}}: To enable real-time inference with Huffman-encoded weights, we parallelize the inherently sequential decoding process. Standard Huffman decoding processes variable-length symbols one at a time, causing latency on edge devices. We address this by segmenting the encoded parameter space while preserving the original weight-tensor structure. Each segment has known boundaries and can be decoded independently, allowing parallel processing across threads without synchronization. To balance workloads, we shuffle multiple segments per thread, mitigating skewed symbol distributions and ensuring consistent decoding times. This approach removes the serial bottleneck, enabling fast, low-latency inference suitable for time-critical applications. Figure \ref{fig:parallel-decoding} illustrates the design.
\vspace{-4mm}
\section{Experiments}
\label{sec:experiments}
\vspace{-4mm}

We evaluated our proposed compression scheme on three edge-based LLMs: \textit{smolLM-1.7B-Instruct} (1.7 billion parameters) \cite{SmolLM-2024}, \textit{phi3-mini-4k-Instruct} (3.8 billion parameters) \cite{phi3-2024} and \textit{mistral-7B-Instruct} (7 billion parameters) \cite{jiang2023mistral7b}. Our code and models are available online \cite{our_code}. Table \ref{tab:pplbenchmark} shows their baseline \texttt{fp16} sizes and subsequent sizes after quantization and Huffman encoding. Figure \ref{fig:plots} visually confirms the Gaussian distribution of the quantized weights. \\
\noindent\textbf{Storage Reduction}: Storage reduction occurs in two steps. First, models are quantized, e.g., \textit{phi3-mini-4k-Instruct} shrinks from 7.2 GB to 3.6 GB (\texttt{uint8}) and 1.8 GB (\texttt{uint4}). Next, Huffman compression further lowers effective bit-widths to 5.58 (\texttt{uint8}) and 1.39 (\texttt{uint4}) bits. Table \ref{tab:pplbenchmark} shows similar results for \textit{smolLM-1.7B-Instruct} and \textit{mistral-7B-Instruct}. Moving from 8-bit (256 symbols) to 4-bit (16 symbols) creates a “bucketing” effect, where frequent central values dominate, reducing randomness or entropy. Our tensor-level approach, with its ability to sharpen the quantized weight distribution, is even more effective in improving compressibility at lower bitwidths compared to block level methods. We boost compressibility by 7$\times$ (8-bit) and 11.3$\times$ (4-bit) over state-of-the-art methods, enabling greater overall compression (Table \ref{tab:compression}, Fig. \ref{fig:plots}). 
\begin{table}[b!]
\vspace{-4mm}
	\centering
	\resizebox{\columnwidth}{!}{
		\begin{tabular}{||c|c|c|c||}
			\hline\hline
			\multirow{2}{*}{\textsc{Task}} & \multirow{2}{*}{\textsc{Encoding}} & \textsc{Latency w/o Huffman} & \textsc{Latency w/ Huffman} \\
			& & (sec) & (sec) \\
			\hline\hline
			pre-fill & \multirow{4}{*}{\texttt{uint8}}  & $27.10$ & $\mathbf{23.17}$ \\
			\cline{1-1}\cline{3-4}
			token generation &  & $0.083$  & $\mathbf{0.063}$ \\
			\cline{1-1}\cline{3-4}
			parallel decoding &  & - & $6.66$ \\
			\cline{1-1}\cline{3-4}
			first token latency & & $\mathbf{27.18}$ & $29.89$ \\
			\hline\hline
			pre-fill & \multirow{4}{*}{\texttt{uint4}} & $9.69$ & $\mathbf{8.34}$ \\
			\cline{1-1}\cline{3-4}
			token generation &  & $0.062$  & $\mathbf{0.025}$ \\
			\cline{1-1}\cline{3-4}
			parallel decoding &  & - & $1.66$ \\
			\cline{1-1}\cline{3-4}
			first token latency & & $\mathbf{9.75}$ & $10.03$ \\
			\hline\hline
	\end{tabular}}
    \vspace{-6mm}
	\medskip
	\caption{\footnotesize Latency breakdown for the \textit{phi3-mini-4k} model on an NVIDIA Jetson P3450 across different quantization formats (\texttt{uint8} and \texttt{uint4}) with and without Huffman compression.}
	\label{tab:latency}
    \vspace{-2mm}
\end{table}

\noindent\textbf{Performance}: We tested our compressed models on a range of tasks to evaluate their performance under low-precision quantization. Specifically, we considered text generation on the \textsc{Wikitext2} benchmark, common sense reasoning on the 5-shot \textsc{HellaSwag} benchmark, and arithmetic reasoning on the 8-shot \textsc{GSM8k CoT} benchmark. Table \ref{tab:pplbenchmark} summarizes the results for three model families—\textit{smolLM-1.7B-Instruct}, \textit{phi3-mini-4k-Instruct} \textit{mistral-7B-Instruct} -- evaluated across \texttt{fp16}, \texttt{uint8}, and \texttt{uint4} precision levels.




\noindent\textbf{Hardware Efficiency \& Parallel Decoding}: Table \ref{tab:latency} shows latency measurements for the \textit{phi3-mini-4k-Instruct} model ($3.8$B parameters) evaluated on an \textsc{NVIDIA Jetson P3450} with four CPU threads for parallel decoding. The device features a quad-core ARM Cortex-A57 CPU ($1.43$ GHz), $4$ GB LPDDR4 memory (25.6 GB/s), and a 128-core Maxwell GPU. NEON SIMD support \cite{arm-cortex-a57} enables efficient bit-level parallelism, accelerating low-precision Huffman decoding for \texttt{uint4} and \texttt{uint8} models. Parallel decoding is highly efficient, especially for \texttt{uint4}, where decoding takes only $1.66$ s versus $9.69$ s for pre-fill and $9.75$ s for the first output token. Even for \texttt{uint8}, decoding completes in $6.66$ s, a small fraction of total inference time. Since decoding occurs only once per sequence, its cost is amortized over many tokens, making the overhead negligible. This enables fast, low-power, and latency-friendly inference on edge benchmarks like \textsc{Wikitext2}, \textsc{HellaSwag}, and \textsc{GSM8k}, demonstrating the practicality of our compressed models for edge deployments.

\noindent\textbf{Latency Reduction from Huffman Coding}: Table \ref{tab:latency} highlights the performance gains from applying Huffman coding to quantized \textit{phi3-mini-4k} weights on the \textsc{NVIDIA Jetson P3450}. For \texttt{uint8}, Huffman coding reduces the effective bit-width to $5.58$ bits, cutting pre-fill time by $14.5\%$ and token generation latency by $31.9\%$. For \texttt{uint4}, it compresses to $1.39$ bits, yielding $13.3\%$ faster pre-fill and a $146.6\%$ improvement in token generation. These gains stem from reduced memory transfers during the memory-bandwidth-limited decoding phase \cite{pope2023efficiently, shazeer2019fast}. When 8-bit weights compress to $5.58$ bits, $30\%$ less data is moved, yielding a theoretical $1.43\times$ speedup, closely matched by our measured $1.32\times$. Pre-fill improvements are smaller since this phase is compute-bound \cite{kwon2023efficient}. Huffman decoding is done once per sequence, adding negligible overhead when amortized over many tokens. Although GPUs lack native support for fractional bit-widths, optimized CUDA kernels efficiently pack/unpack values \cite{fang2023turbomixer}, keeping compute precision fixed and focusing on memory transfer optimization.

\vspace{-4mm}
\section{Conclusions}
\label{sec:conc}
\vspace{-4mm}
We present EntroLLM, a framework combining mixed quantization, Huffman compression, and parallel decoding to enable efficient LLM deployment on edge devices. Our method reduces average bit-width to 1.39 for 4-bit weights, improving downstream entropy coding by $7\times$–$11.3\times$ over state-of-the-art techniques. Parallel decoding keeps decompression overhead negligible, even on low-power hardware like the \textsc{NVIDIA Jetson P3450}. Experiments show significant latency and storage gains, enabling real-time, low-power inference for applications such as robotics and IoT. Future work will explore adaptive entropy coding and hardware-aware optimizations.

\bibliographystyle{IEEEbib}
\bibliography{main}

\end{document}